# ABC-SG: A New Artificial Bee Colony Algorithm-Based Distance of Sequential Data Using Sigma Grams

**Muhammad Marwan Muhammad Fuad**
Department of Electronics and Telecommunications
Norwegian University of Science and Technology (NTNU)
NO-7491 Trondheim, Norway
marwan.fuad@iet.ntnu.no

**Abstract**
The problem of similarity search is one of the main problems in computer science. This problem has many applications in text-retrieval, web search, computational biology, bioinformatics and others. Similarity between two data objects can be depicted using a similarity measure or a distance metric. There are numerous distance metrics in the literature, some are used for a particular data type, and others are more general. In this paper we present a new distance metric for sequential data which is based on the sum of n-grams. The novelty of our distance is that these n-grams are weighted using artificial bee colony; a recent optimization algorithm based on the collective intelligence of a swarm of bees on their search for nectar. This algorithm has been used in optimizing a large number of numerical problems. We validate the new distance experimentally.

## 1 Introduction

Similarity search is one of the fundamental problems in computer science. It has many applications in text, video and image retrieval, pattern recognition, bioinformatics, web search, fingerprint databases, and many others. In this problem a pattern is given and the algorithm searches the database, or the web, to return all or most, depending on whether the search is exact or approximate, of the data objects that are "close" to that pattern according to some semantics of closeness. This closeness between two data objects is depicted using a principal concept which is the similarity measure or its stronger form; the distance metric.

Of the different paradigms proposed to manage the similarity search problem, the metric model with its properties (reflexivity, non-negativity, symmetry, triangle inequality) stands out as one that is applicable to different data types. The distance metric on which the metric model is based is a strong mathematical tool which helps the researchers build different data structures specific to metric spaces. Other techniques, such as the pivot technique, are based on the triangle inequality; one of the axioms of the metric model. All these advantages of this model make of it a rich field of research in information retrieval.

The main distance used to compare two strings is the *Edit Distance* (ED) presented by Wagner and Fischer (1974), it is also called the *Levenshtein distance,* and it is defined as the minimum number of delete, insert, and change operations needed to transform string *S* into string *T*. As mentioned above, this distance is the main distance used to compare two strings. However, this distance has its limitations because it considers local similarity only.

Muhammad Fuad and Marteau (2008a) (2008b) presented a new distance metric; *The Extended Edit Distanc* (EED), which they applied to symbolically represented time series. Unlike ED, EED considers a global level of similarity in additional to the local one presented by ED. EED is based on the idea of computing the frequencies of common characters between two strings. Later, Muhammad Fuad and Marteau (2008c) presented another distance, MREED, which computes the frequencies of common bi-grams in addition to common characters. However, the parameters used in these two distances (one in EED and two in MREED) were defined using very basic heuristics which, on the one hand, substantially limited the search space (it was limited to 5 values only for each parameter), and on the other hand, using such basic heuristics makes it practically impossible to extend this distance beyond that of bi-grams because training time is very long even in the case of bi-grams where two parameters only are used

In this paper we propose a new general distance metric that applies to strings. We call it the *Artificial Bee Colony-Sigma Gram Distance* (ABC-SG). This distance is based on computing the sigma grams. The particularity of this distance is that it uses the artificial bee colony algorithm to set its parameters.

The rest of this paper is organized as follows: Section 2 is a background section, In Section 3 we present the new distance and we validate its performance in Section 4, we conclude this paper in Section 5 with some perspectives.

## 2 Background

Muhammad Fuad and Marteau (2008a) (2008b) presented the *Extended Edit Distance* (EED) which is defined as follows:

This work was carried out during the tenure of an ERCIM "Alain Bensoussan" Fellowship Programme. This Programme is supported by the Marie-Curie Co-funding of Regional, National and International Programmes (COFUND) of the European Commission.

Let $\Sigma$ be a finite alphabet, and let $\Sigma^*$ be the set of strings on $\Sigma$. Let $f_a^{(S)}, f_a^{(T)}$ be the frequency of the character $a$ in $S$ and $T$, respectively. Where $S$, $T$ are two strings in $\Sigma^*$. EED is defined as:

$$EED(S,T) = ED(S,T) + \lambda \left[ |S| + |T| - 2 \sum_{a \in \Sigma} min\left(f_a^{(S)}, f_a^{(T)}\right) \right] \quad (1)$$

Where $|S|$, $|T|$ are the lengths of the two strings $S, T$ respectively, and where $\lambda \geq 0$ ($\lambda \in R$). $\lambda$ is called the co-occurrence frequency factor.

EED is based on the intuition that the ED distance does not take into account whether the change operation used a character that is more "familiar" to the two strings or not, because ED considers a local level of similarity only, while EED adds to this local level of similarity a global one. This modification makes EED more intuitive as shown by Muhammad Fuad and Marteau (2008a) (2008b).

Muhammad Fuad and Marteau (2008c) also showed that EED is a distance metric (symmetry, identity, triangle inequality). Search in metric spaces has many advantages, the most famous of which is that a single indexing structure can be applied to several kinds of queries and data types that are so different in nature. This is mainly important in establishing unifying models for the search problem that are independent of the data type. This makes metric spaces a solid structure that is able to deal with several data types as mentioned by Zezula et al. (2005).

## 3 The Artificial Bee Colony Sigma Gram Distance (ABC-SG)

### 3.1 Definition-The Number of Distinct n-Grams (NDnG)

Given two strings $S$, $T$. The number of distinct *n-grams* (substrings of length *n*) that the two strings $S$ and $T$ contain is defined as:

$$ND_nG(S,T) = \left| \{n-gram(S)\} \cup \{n-gram(T)\} \right| \quad (2)$$

where $n-gram(\ )$ is the set of *n-grams* that a string consists of.

**Example :**

Given the following strings: $R = oxygen$, $S = exogen$, $T = emolen$. The sets of n-grams for these strings are given by:

| n | R | S | T |
|---|---|---|---|
| 1 | o, x, y, g, e, n | e, x, o, g, e, n | e, m, o, l, e, n |
| 2 | ox, xy, yg, ge, en | ex, xo, og, ge, en | em, mo, ol, le, en |
| 3 | oxy, xyg, yge, gen | exo, xog, oge, gen | emo, mol, ole, len |
| 4 | oxyg, xyge, ygen | exog, xoge, ogen | emol, mole, olen |
| 5 | oxyge, xygen | exoge, xogen | emole, molen |
| 6 | oxygen | exogen | emolen |

Comparing $ND_nG(S,R)$, and $ND_nG(S,T)$ gives:

| n | NDnG(S,R) | NDnG(S,T) |
|---|---|---|
| 1 | 5 | 4 |
| 2 | 2 | 1 |
| 3 | 1 | 0 |
| 4 | 0 | 0 |
| 5 | 0 | 0 |
| 6 | 0 | 0 |
| $\sum_{n=1}^{6} ND_nG$ | 8 | 5 |

The above comparison shows a greater similarity between $S$ and $R$ than between $S$ and $T$, which is intuitive. But if we compute the edit distance we get: $ED(S,R) = ED(S,T) = 2$.

Muhammad Fuad and Marteau (2008a) (2008b) showed how EED, which considers the frequencies of characters, can capture this intuitive similarity that ED can not capture.

Although EED has advantages over ED as shown by Muhammad Fuad and Marteau (2008a), the way parameter $\lambda$ is defined remains problematic. On the one hand, the search space is very limited, on the other hand, generalizing EED to use higher order frequencies of common grams using the same basic heuristics to define the different parameters makes the parameter defining process, for the different grams, inefficient and yet limited to very small regions in the search space.

In the following we present a generalizing of EED which uses an artificial bee colony based approach to determine the different parameters. This makes the search process more efficient and effective.

### 3.2 ABC-SG

Let $\Sigma$ be a finite alphabet, and let $\Sigma^*$ be the set of strings on $\Sigma$. Given *n*, let $f_{a_n}^{(S)}$ be the frequency of the *n-gram* $a_n$ in $S$, and $f_{a_n}^{(T)}$ be the frequency of the *n-gram* $a_n$ in $T$, where $S$, $T$ are two strings in $\Sigma^*$. Let **N** be the set of integers, and $\mathbf{N}^+$ the set of positive integers.

For notation convenience, we define the function:

$$g : \mathbf{N}^+ \times \Sigma^* \to \mathbf{N}$$

$$g(n,S) = n \quad \text{if} \quad 1 \leq n \leq |S|$$

$$g(n,S) = |S| + 1 \quad \text{if} \quad |S| < n$$

The ABC-SG distance between $S$ and $T$ is thus defined as:

$$ABC-SG(S,T) = \sum_{n=1}^{max(|S|,|T|)} \lambda_n \cdot \begin{bmatrix} |S| + |T| - g(n,S) - g(n,T) \\ + 2 - 2 \cdot \sum_{a_n \in A^n} min\left(f_{a_n}^{(S)}, f_{a_n}^{(T)}\right) \end{bmatrix} \quad (3)$$

where $|S|, |T|$ are the lengths of the two strings $S$, $T$ respectively, and where $\lambda_n \in \mathbf{R}^+ \cup \{0\}$.

ABC-SG is based on the same concept of familiarity on which EED is based, but this concept is extended to the familiarity of n-grams instead of that of single characters.

It is important to notice that ABC-SG is actually the generalization of both EED (Muhammad Fuad and Marteau 2008a), (Muhammad Fuad and Marteau 2008b) and MREED (Muhammad Fuad and Marteau 2008c), so it includes the same advantages that these two distances have.

ABC-SG is proved to be a distance metric. For space limitations, the proof is not presented here. However, the proof is an extension of the proof presented by Muhammad Fuad and Marteau (2008a) (2008b).

As indicated earlier, the parameters $\lambda_n$ are determined using the artificial bee colony algorithm.

### 3.3 Artificial Bee Colony

*Bee-inspired* optimization is a family of optimization algorithms that emerged from a larger family which is *swarm intelligence*. Baykasoğlu, Özbakır and Tapkan (2007) classify the behavioral characteristics of bee-based algorithms into three categories: foraging behaviors, marriage behaviors, and queen bee concept. One of the foraging behavior-based algorithms is *Artificial Bee Colony* (ABC) which was introduced by Karaboga (2005). In ABC each food source represents a potential solution to the optimization problem at hand and the quality of the food represents the value of the objective function to be optimized. Artificial bees explore and exploit the search space. These bees communicate and share information about the location and quality of food sources. This exchange of information takes place in the dancing area in the hive by performing a *waggle dance*.

In ABC there are three kinds of bees:

**Employed bees:** These are the bees that search in the neighborhood of a food source. They perform a dance with a probability that is proportional to the quality of the food source.

**Onlooker bees:** These bees are found on the dance floor. They watch the dances of the employed bees and place themselves on the most profitable food source.

**Scouts:** These bees explore the search space randomly.

As mentioned by Parpinelli, Benitez, and Lopes (2010), the balance between exploration and exploitation is maintained in ABC algorithm by combining local search methods, carried out by the employed and the onlooker bees, with global search methods, carried out by the scouts.

There are several variations of the ABC algorithm. In the following we present the standard ABC introduced by Karaboga and Basturk (2007a) (2007b), and by Diwold Beekman, and Middendorf (2010). The first step of ABC is generating a randomly distributed population size (pop_size) of food sources which correspond to potential solutions. Each solution $\vec{x}_i, i \in \{1,..,pop\_size\}$ is a vector whose dimension is (nr_par) which is equal to the number of parameters of the function $f$ to be optimized. The population is subject to change for a number of cycles (nr_cycles). In each cycle every employed bee perturbs the current solution using a local search procedure. The perturbation produces a new solution:

$$\vec{x}_i^* = \vec{x}_i + rand(-1,1)(\vec{x}_i - \vec{x}_k) \quad , i \neq k \quad (4)$$

The above relation is not applied to all parameters but only to a certain number of them. The parameters to be altered are chosen randomly. The algorithm uses a greedy selection to decide if the new solution should be kept or discarded, i.e. :

---

**Algorithm1** Artificial Bee Colony (ABC)

**Require** `pop_size, nr_par, nr_cycles, max_nr.`

```
1:  Initialize x_i
2:  for cycle=1 to nr_cycles do
3:      for all employed bees do
4:          x_i* = x_i + rand(-1,1)(x_i - x_k) ,i ≠ k
5:          if f(x_i*) < f(x_i) then  x_i ← x_i*
6:      end for
7:      calculate  p_i = f(x_i) / Σ_{k=1}^{pop_size} f(x_k)
8:      for all onlooker bees do
4:          x_i* = x_i + rand(-1,1)(x_i - x_k) ,i ≠ k
5:          if f(x_i*) < f(x_i) then  x_i ← x_i*
6:      end for
11:     if nr_of_trials== max_nr then
            abandon current solution
12: end for
```

---

**Figure 1. The Artificial Bee Colony Algorithm**

$$\vec{x}_i = \begin{cases} \vec{x}_i^* & \text{if} \quad f(\vec{x}_i^*) < f(\vec{x}_i) \\ \vec{x}_i & \text{otherwise} \end{cases} \quad (5)$$

After all employed bees have modified their positions the onlooker bees choose one of the current solutions depending on a probability that corresponds to the fitness value of that solution according to the following rule:

$$p_i = \frac{f(\vec{x}_i)}{\sum_{k=1}^{pop\_size} f(\vec{x}_k)} \quad (6)$$

After that the onlooker bees try to improve the solution using the same mechanism that was described in (4). The number of trials the algorithm attempts to improves the same solution is limited by a maximum number (*max_nr*) after which the solution is abandoned and the bees employed by that food source become scouts. The abandoned solution is replaced by a new solution found by the scouts. Figure 1 outlines the ABC algorithm.

## 4 Performance Evaluation

We tested the new distance ABC-SG on symbolically represented time series. However, we think that ABC-SG is more appropriate for other sequential data types such as those encountered in bioinformatics and text mining.

Time series data are normally numeric, but there are different methods to transform them to symbolic data. The most important symbolic representation method of time series is the *Symbolic Aggregate Approximation* (SAX) introduced by Lin, Keogh, Lonardi, and Chiu (2003). SAX is based on an assumption that normalized time series have Gaussian distribution, so by determining the breakpoints that correspond to a particular alphabet size, one can obtain equal-sized areas under the Gaussian curve. SAX is applied as follows:

1-The time series are normalized.

2-The dimensionality of the time series is reduced using PAA; a representation method presented independently by Keogh, Chakrabarti, Pazzani, and Mehrotra (2000) and by Yi and Faloutsos (2000).

3-The PAA representation of the time series is discretized by determining the number and locations of the breakpoints (The number of the breakpoints is chosen by the user). Their locations are determined, as mentioned above, using Gaussian lookup tables. The interval between two successive breakpoints is assigned to a symbol of the alphabet, and each segment of PAA that lies within that interval is discretized by that symbol.

The last step of SAX is using the following similarity measure:

$$MINDIST(\hat{S}, \hat{R}) \equiv \sqrt{\frac{n}{N}} \sqrt{\sum_{i=1}^{N} (dist(\hat{s}_i, \hat{r}_i))^2} \quad (7)$$

Where $n$ is the length of the original time series, $N$ is the length of the strings (the number of the segments), $\hat{S}$ and $\hat{R}$ are the symbolic representations of the two time series $S$ and $R$, respectively, and where the function $dist(\ )$ is implemented by using the appropriate lookup table.

We also need to mention that the similarity measure used in PAA is:

$$d(S, R) = \sqrt{\frac{n}{N}} \sqrt{\sum_{i=1}^{N} (\overline{s}_i - \overline{r}_i)^2} \quad (8)$$

It is important to mention that MINDIST is not a distance metric (because it violates the axioms of distance metric) but a similarity measure.

We tested our new distance ABC-SG on a time series classification task based on the first nearest-neighbor (1-NN) rule using leaving-one-out cross validation. This means that every time series is compared to the other time series in the dataset. If the 1-NN does not belong to the same class, the error counter is incremented by 1.

We conducted experiments using datasets of different sizes and dimensions available at UCR of Keogh, Zhu, Hu, Hao, Xi, Wei, and Ratanamahatana (2011). This archive makes up between 90% and 100% of all publicly available, labeled time series data sets in the world, as mentioned by Ding, Trajcevski, Scheuermann, Wang, and Keogh (2008).

As indicated earlier, we tested ABC-SG on symbolically represented time series This means that the time series were transformed to symbolic sequences using the first three step of SAX presented earlier in this section, but instead of using MINDIST given in relation (7), we use our distance ABC-SG. The parameters $\lambda_n$ in the definition of ABC-SG (relation (3)) are defined using ABC. This means, for each value of the alphabet size we formulate an artificial bee colony optimization problem where the fitness function is the classification error and the parameters of the optimization problem are $\lambda_n$. Theoretically $n$ can take any value that does not exceed that of the shortest string of the two strings $S$, $T$. However, in the experiments we conducted we tested the new distance for $n \in \{1,2,3\}$ because these are the values of interest for time series. Notice that ABC-SG can be applied to strings of different lengths, which is one of its advantages since most similarity measures in time series mining are applied only to time series of the same length.

Concerning the control parameters of the ABC we used, the population size (the number of food sources) *pop_size* was 20. The number of cycles *nr_cycles* was set to 20. The number of trials of a certain food source *max_nr* was set to 10. The number of parameters *nr_par*, as mentioned earlier, was tested for $n \in \{1,2,3\}$. As for $\lambda_n$, their values are in fact unconstrained, but for simplicity we optimized them in the interval $[0,2]$. Table 1 summarizes the symbols used in the experiments together with their corresponding values.

| | | |
|---|---|---|
| *pop_size* | Population size | 20 |
| *nr_cycles* | Number of cycles | 20 |
| *max_nr* | Number of trials | 10 |
| *nr_par* | Number of parameters | {1,2,3} |

**Table 1. The symbol table of ABC together with the corresponding values used in the experiments**

For each dataset we use ABC on the training datasets to get the vector $\lambda_n$ that minimizes the classification error on this training dataset, and then we use these optimal values of $\lambda_n$ on the corresponding testing dataset to get the final classification error for each dataset.

We compared ABC-SG with dynamic time warping (DTW). DTW is a similarity measure that has been developed by the speech recognition community and later was used by Berndt, and Clifford (1994) on time series. DTW is an algorithm to find the optimal path through a matrix of points representing possible time alignments between the signals. Guo and Siegelmann (2004) state that the optimal alignment can be efficiently calculated via dynamic programming.

The dynamic time warping between the two time series $S = \{s_1, s_2, ..., s_n\}$, $R = \{r_1, r_2, ..., r_m\}$ is defined as follows:

$$DTW(i,j) = d(i,j) + min\begin{cases} DTW(i, j-1) \\ DTW(i-1, j) \\ DTW(i-1, j-1) \end{cases} \quad (9)$$

where $1 \leq i \leq n, 1 \leq j \leq m$.

We chose to compare ABC-SG with DTW because DTW is known to give very good results in several time series data mining tasks such as classification and clustering. Another reason for choosing DTW is because it is applicable to time series of different lengths, which is the case with ABC-SG. However, ABC-SG has a complexity of $O(n^2)$, while that of ABC-SG is $O(N^2)$ ($N = n/4$ for compression ratio 1:4; the compression ratio usually used with SAX). So as we can see, ABC-SG has a much lower complexity than that of DTW. Another advantage that ABC-SG has over DTW is that ABC-SG is a distance metric while DTW is a similarity measure because it violates the triangle inequality.

It is important to mention that DTW is applied to the original time series and not to their symbolic representation.

In Table 2 we present some of the results we obtained for alphabet size equal to 3, 10, and 20, respectively.

As we can see from the results, the classification errors of ABC-SG are quite comparable to those of DTW despite the difference in complexity. In fact, in the majority of cases, ABC-SG even outperformed DTW. The results of other datasets in the archive were similar.

**Beef**

| | ABC-SG | | | DTW |
|---|---|---|---|---|
| | n=1 | n=2 | n=3 | |
| α* = 3 | 0.567 | 0.567 | 0.567 | |
| α = 10 | 0.5 | 0.5 | 0.467 | 0.5 |
| α = 20 | 0.333 | 0.367 | 0.367 | |

(*: α is the alphabet size)

**ECG**

| | ABC-SG | | | DTW |
|---|---|---|---|---|
| | n=1 | n=2 | n=3 | |
| α = 3 | 0.18 | 0.21 | 0.22 | |
| α = 10 | 0.2 | 0.22 | 0.22 | 0.23 |
| α = 20 | 0.23 | 0.22 | 0.25 | |

**FaceFour**

| | ABC-SG | | | DTW |
|---|---|---|---|---|
| | n=1 | n=2 | n=3 | |
| α = 3 | 0.057 | 0.057 | 0.057 | |
| α = 10 | 0.045 | 0.057 | 0.068 | 0.170 |
| α = 20 | 0.09 | 0.102 | 0.102 | |

**OSULeaf**

| | ABC-SG | | | DTW |
|---|---|---|---|---|
| | n=1 | n=2 | n=3 | |
| α = 3 | 0.351 | 0.343 | 0.331 | |
| α = 10 | 0.298 | 0.306 | 0.298 | 0.409 |
| α = 20 | 0.322 | 0.330 | 0.330 | |

**Table 2. Comparison between the classification error of ABC-SG and DTW for different values of the alphabet size and for different n-grams.**

Another interesting remark which makes this distance meaningful is that we did not witness any correlation between the number of grams used and the classification error, which makes sense since ABC-SG, as mentioned in Section 3, is based on the concept of familiarity of n-grams between the two strings, and this familiarity is not related to the length of the n-gram.

Finally, in Table 3 we present, for reproducibility purposes, the values of $\lambda_n$ obtained on the training datasets. As indicated earlier, when applying these values to the corresponding testing datasets we obtain the final classification errors presented in Table 2

| Dataset | Alphabet size | n-gram | $\lambda_n$ |
|---|---|---|---|
| Beef | 3 | 1 | [0.14897] |
| | | 2 | [0.059199  0.64354] |
| | | 3 | [0.18031  0.51181  0.0013076] |
| | 10 | 1 | [1.2358] |
| | | 2 | [0.97218  0.38007] |
| | | 3 | [0.80256  0.79378  0.19502] |
| | 20 | 1 | [0.19036] |
| | | 2 | [0.91433  0.18418] |
| | | 3 | [0.76518  0.0053491  0.067656] |
| ECG | 3 | 1 | [0.80394] |
| | | 2 | [0.82499  0.25791] |
| | | 3 | [0.66397  0.58473  0.54427] |
| | 10 | 1 | [0.74811] |
| | | 2 | [0.1243  0.55862] |
| | | 3 | [0.11591  1.7659  0.81072] |
| | 20 | 1 | [0.076999] |
| | | 2 | [0.022884  0.84058] |
| | | 3 | [0.027087  0.74227  0.439] |
| FaceFour | 3 | 1 | [0.22144] |
| | | 2 | [0.21865  0.26487] |
| | | 3 | [0.22744  0.13813  0.031161] |
| | 10 | 1 | [0.17061] |
| | | 2 | [0.048042  0.10718] |
| | | 3 | [0.054286  0.076333  0.11382] |
| | 20 | 1 | [0.030194] |
| | | 2 | [0.22075  0.10383] |
| | | 3 | [0.26036  0.053799  0.024754] |
| OSULeaf | 3 | 1 | [0.53548] |
| | | 2 | [0.66298  0.060555] |
| | | 3 | [0.65041  0.039146  0.052083] |
| | 10 | 1 | [0.54977] |
| | | 2 | [0.51024  0.020506] |
| | | 3 | [0.38619  0.039477  0.19883] |
| | 20 | 1 | [0.34882] |
| | | 2 | [0.28517  0.025615] |
| | | 3 | [0.073991  0.020775  0.063922] |

**Table 3. The optimal values of $\lambda_n$ obtained by applying ABC to the training datasets.**

## 5 Conclusion

In this paper we presented a new distance metric; ABC-SG, which is applied to strings. This distance considers the frequencies of n-grams, which adds a global level of similarity, in addition to the local one. The particularity of this distance is that it uses the artificial bee colony algorithm to determine the values of its parameters. We tested the new distance and we compared it to a very competitive similarity measure; DTW, on a time series classification task. We showed that our distance ABC-SG gave better results in most cases, despite the difference in complexity.

In order to represent the time series symbolically, we had to use SAX because this is the most widely used symbolic representation method of time series. Nonetheless, a representation technique prepared specifically for ABC-SG may even give better results.

Although we used ABC as an optimization algorithm to set the parameters of the new distance, we think other stochastic and bio-inspired optimization algorithms can also be used with the new distance.